\documentclass[10pt,twocolumn,letterpaper]{article}

\usepackage[pagenumbers]{iccv} % To force page numbers, e.g. for an arXiv version
\usepackage{algorithm}
\usepackage{algorithmic}
\usepackage{bm}
\usepackage{amsmath}
\usepackage{amssymb}
\usepackage{pifont}
\usepackage{graphicx}
\usepackage{textcomp}
\definecolor{iccvblue}{rgb}{0.21,0.49,0.74}
\usepackage[pagebackref,breaklinks,colorlinks,allcolors=iccvblue]{hyperref}

\title{GarmentX: Autoregressive Parametric Representations for High-Fidelity \\ 3D Garment Generation}

\def\authorBlock{
    Jingfeng Guo\textsuperscript{1}\textsuperscript{*} \quad 
    Jinnan Chen\textsuperscript{2}\textsuperscript{*} \quad
    Weikai Chen\textsuperscript{3}\textsuperscript{\textdagger} \quad
    Zhenyu Sun\textsuperscript{1} \\
    Lanjiong Li\textsuperscript{4} \quad
    Baozhu Zhao\textsuperscript{1} \quad
    Lingting Zhu\textsuperscript{5}  \quad
    Xin Wang\textsuperscript{3} \quad
    Qi Liu\textsuperscript{1}\textsuperscript{\textdagger} \\
    
    \textsuperscript{1}South China University of Technology  \quad 
    \textsuperscript{2}National University of Singapore  \\
    \textsuperscript{3}LIGHTSPEED \quad
    \textsuperscript{4}The Hong Kong University of Science and Technology (Guangzhou) \\
    \textsuperscript{5}The University of Hong Kong \\
}
\newcommand{\methodName}{GarmentX}

\begin{document}

\twocolumn[{%
\renewcommand\twocolumn[1][]{#1}%
\maketitle

\begin{center}
    \vspace{-4\baselineskip}
    \author{\authorBlock}
    % Project Page: \url{https://DRiVEAvatar.github.io/} \\
\end{center}

\begin{center}
    \centering
    \includegraphics[width=\textwidth]{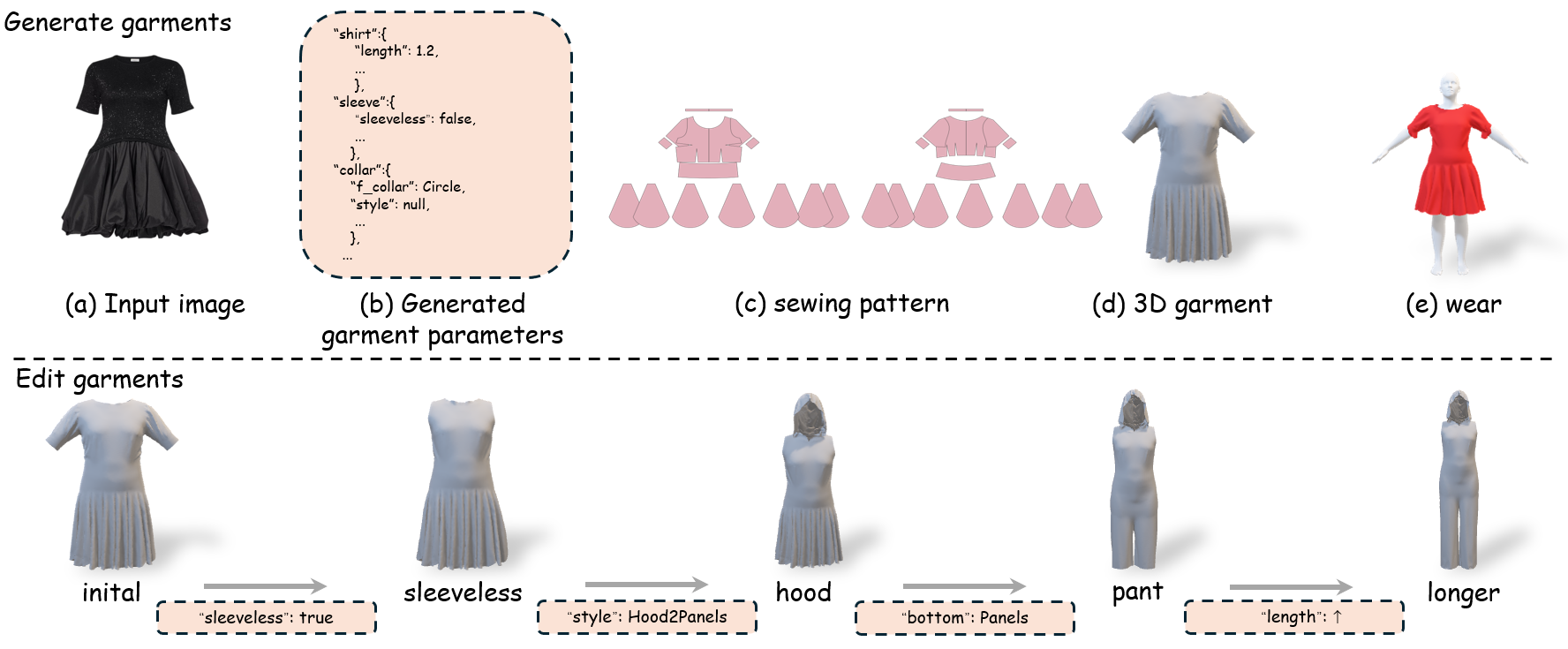}
    \vspace{-3mm}
    \captionof{figure}{
    \methodName{} is an image-guided 3D garment generator that produces editable garment parameters, decodes valid sewing patterns, and simulates wearable 3D garments. By adjusting garment parameters, users can easily modify the shape, style, and even category of the garments. The generated garments are diverse, high-fidelity, and physically plausible.}
    \label{fig: overall}
\end{center}
}]

\renewcommand{\thefootnote}{} % 清除脚注编号
\footnotetext{\textsuperscript{*} Indicates Equal Contribution. \textsuperscript{\textdagger} Indicates Corresponding Author.}
\renewcommand{\thefootnote}{\arabic{footnote}} % 恢复脚注编号

\begin{abstract}
This work presents \methodName{}, a novel framework for generating diverse, high-fidelity, and wearable 3D garments from a single input image. Traditional garment reconstruction methods directly predict 2D pattern edges and their connectivity, an overly unconstrained approach that often leads to severe self-intersections and physically implausible garment structures. In contrast, \methodName{} introduces a structured and editable parametric representation compatible with GarmentCode, ensuring that the decoded sewing patterns always form valid, simulation-ready 3D garments while allowing for intuitive modifications of garment shape and style.
To achieve this, we employ a masked autoregressive model that sequentially predicts garment parameters, leveraging autoregressive modeling for structured generation while mitigating inconsistencies in direct pattern prediction. Additionally, we introduce \methodName{} dataset, a large-scale dataset of 378,682 garment parameter-image pairs, constructed through an automatic data generation pipeline that synthesizes diverse and high-quality garment images conditioned on parametric garment representations. Through integrating our method with \methodName{} dataset, we achieve state-of-the-art performance in geometric fidelity and input image alignment, significantly outperforming prior approaches. We will release \methodName{} dataset upon publication.

% Unlike previous pattern-based approaches that directly predict sewing patterns through each panel's rotational matrices, translation matrices, vertex coordinates, edge information, and stitch information, which frequently result in self-intersections, undetectable panels, or incorrect stitching relation, leading to simulation failures in complex or out-of-distribution cases. Inspired by physical garment manufacturing workflows, our framework introduces a masked autoregressive transformer to generate editable garment parameters, combined with GarmentCode, ensuring the production of reasonable sewing pattern panels and stitch information, thereby avoiding erroneous simulations and achieving more robust and controllable generation. Leveraging 378,682 garment parameter-image pairs generated via innovative automatic data pipeline, our framework demonstrates exceptional performance in producing 3D garments that closely align with input images. Experimental results validate the effectiveness of the proposed framework, achieving state-of-the-art performance in image-to-garment generation with robust generalization capabilities, while ensuring user-friendly editability for practical applications like 3D virtual try-on and digital content creation.
\end{abstract}

\section{Introduction}

3D Garments play a crucial role in digital humans~\cite{GauHuman,hu2025humangif,wang2023disentangled,chen2024generalizable,chen2024dihur,wang2025fresa,yi2025,zhaung24,xu2024highqualityhumanimage,guo2025vid2avatarpro,li2025etch}, and virtual try-on applications~\cite{xie2024dreamvton,vton24,Dong_2019_ICCV,Dong_2019_ICCV2}, serving as essential assets in film VFX, game development, and AR/VR experiences. Recent 3D generation approaches~\cite{li2025triposg,lu2024geal,zhu2025muma,zhao2024di,hu2024x,zhang2024lagem} have provided valuable guidance for transforming how digital clothing items can be created with high-fidelity and physically plausible results. Traditional 3D garment creation pipelines primarily rely on two approaches: (1) manual modeling using professional software like CLO3D~\cite{clo3d} and (2) high-fidelity 3D scanning system 3dMD \cite{3dMD} and Artec3D \cite{Artec3D} that capture garment geometry through multi-camera photogrammetry. However, both methods require either specialized expertise or costly hardware, creating prohibitive barriers for large-scale 3D garment production.

% 3D garments, represented as polygonal mesh data, serve as ubiquitous assets across entertainment and digital content industries including film VFX, game development, and VR/AR experiences. Current creation pipelines predominantly rely on two approaches: 1) skilled artists manual modeling using CAD software like CLO3D \cite{clo3d}, or 2) high-fidelity 3D scanning systems such as 3dMD \cite{3dMD} and Artec3D \cite{Artec3D} that capture garment geometry through multi-camera photogrammetry. This dependency on specialized labor or capital-intensive hardware creates prohibitive barriers for large-scale 3D garment production.

% Recent deep learning-based methods \cite{he2024dresscode,li2024garmentdreamer,liu2024clothedreamer,srivastava2025wordrobe} have explored text-to-garment generation, but text descriptions often only summarize garment categories and fail to capture precise details. For instance, the input prompt "a circle-neck short-sleeve dress" cannot precisely specify attributes such as sleeve length, cuff width, collar depth, or dress length. In contrast, images provide a more comprehensive representation of such details, making them more advantageous for practical applications. Our work specifically targets this complex situation. 

Recently, deep learning methods have been explored for text-to-garment and image-to-garment generation. While text-based methods~\cite{he2024dresscode,li2024garmentdreamer,liu2024clothedreamer,srivastava2025wordrobe} provide high-level control, they struggle to capture fine-grained garment details such as sleeve width, skirt flare, or neckline shape. Image-based methods offer richer details but still face severe limitations in structural accuracy and editability.
The image-to-garment generation direction can be categorized into two types: pattern-based methods \cite{chen2024panelformer,liu2023towards,pietroni2022computational,nakayama2024aipparel,bian2024chatgarment} and deformation-based approaches \cite{sarafianos2024garment3dgen,de2023drapenet,luo2024garverselod}. 
% However, both categories exhibit critical limitations. 
Pattern-based approaches predict sewing pattern edges, stitch connectivity, and 2D panel relationships. However, this formulation is overly unconstrained, often leading to self-intersections, misaligned panels, and physically invalid patterns. 
% Pattern-based methods directly predict each panel's rotation matrix, translation matrix, vertex coordinates, edge information, and stitch information to obtain a sewing pattern, which is then used to simulate 3D garments. However, they often produce self-intersecting panels or incorrect stitching relationships when handling complex cases or garments that fall beyond their training distribution. 
Deformation-based methods rely on predefined 3D garment templates, which are warped to fit the target shape. However, these methods are highly dependent on the chosen template, making them prone to misalignment when dealing with unseen garment styles or out-of-distribution shapes.
% Significant discrepancies between target garments and templates often result in misaligned or incomplete geometries relative to input images. Additionally, their computational intensity renders them impractical for industrial applications. 
Meanwhile, general image-to-3D approaches \cite{xu2024instantmesh,tang2025lgm,xiang2024structured,tochilkin2024triposr} can produce matching results, but these results often contain closed or double-layered structures, making them unwearable.

% Given these limitations, there is an urgent need to develop a method for generating high-fidelity, wearable, simulation-ready, and highly generalizable 3D garments from a single image.

To address these challenges, we introduce \methodName{}, a parametric driven framework for structured, editable, and physically valid 3D garment generation.
Unlike previous approaches that directly predict sewing pattern edges or deform template meshes, \methodName{} represents garments through a structured parametric representation that defines high-level, semantically meaningful attributes. 
By predicting garment parameters instead of raw pattern edges, GarmentX ensures that the generated garments maintain correct stitching relationships, geometric integrity, and simulation robustness. This structured formulation not only enhances the quality of the generated garments but also makes them highly controllable and editable.

Given the unordered nature of garment parameters and the inherent advantages of autoregressive models in visual generation—particularly their superior scalability and generation speed— we leverage a masked autoregressor for garment parameter generation. 
By employing an autoregressive model, we decompose the complex joint distribution modeling into a product of multiple simpler marginal distributions, making generation more tractable. Our autoregressor incorporates a diffusion process to effectively handle the continuous values of garment parameters, enabling more precise and flexible customization. 

To support training and evaluation, we construct \methodName{} dataset, a data collection of 378,682 garment parameter-image pairs, built using an automatic data construction pipeline. Unlike prior datasets, which rely on low-level pattern descriptors, we transform GarmentCodeData's \cite{korosteleva2024garmentcodedata} raw attributes into \methodName{} structured representation, ensuring that garments remain semantically meaningful and editable.
To bridge the domain gap between synthetic rendering and real-world garments, we further employ a canny-conditioned ControlNet model~\cite{zhang2023adding} to synthesize high-quality, photorealistic garment images, improving the robustness of the learned garment representation.
Extensive experiments show that \methodName{} outperforms state-of-the-art methods in geometry fidelity, input-image alignment, and wearability. We summarize our contributions:

% Furthermore, due to the lack of paired garment images and corresponding garment parameters, we propose an automatic data construction pipeline that leverages the GarmentCodeData \cite{korosteleva2024garmentcodedata} dataset and techniques like canny-conditional Stable Diffusion \cite{zhang2023adding,mou2024t2i} or texture generation \cite{zeng2024paint3d,zhang2024dreammat} for garment image generation and rendering. This pipeline enables us to build a large-scale dataset of garment parameters-image pairs to train our model. We summarize our contributions as follows:

\begin{itemize}
    \item We introduce \methodName{}, a novel masked autoregressive framework that models garments through a structured and editable parametric representation, ensuring controllable and physically valid garment generation. 
    % \item We propose a masked autoregressive (MAR) model for garment parameter generation to circumvent the inherent challenges associated with direct sewing pattern prediction.
    % \item We develop an automatic data construction pipeline to obtain large-scale garment parameter-image pairs to train our model.
    \item We build and release \methodName{} dataset, a collection of 378,682 garment parameter-image pairs generated by an automatic data construction pipeline.
    The dataset provides diverse, high-fidelity garment images aligned with parametric descriptions.
    \item We achieve the state-of-the-art results in single-view garment generation on several benchmarks, with particular improvements in geometric fidelity, generalization capability, and input-image alignment.
\end{itemize}

\section{Related work}
% \subsection{3D Garment Creation}
\paragraph{Open Surface and Garment Pattern Generation} Open surface generation has advanced substantially through unsigned distance fields (UDFs)~\cite{yu2023surf,DreamUDF2024}. For clothing, sewing patterns provide a superior approach by precisely representing garment structure without the difficult gradient predictions of UDFs. Pattern-based techniques predict panel and stitch data before 3D simulation. Notable advances include NeuralTailor's \cite{korosteleva2022neuraltailor} LSTM-based pattern reconstruction, transformer architectures from SewFormer \cite{liu2023towards} and PanelFormer \cite{chen2024panelformer} for image-to-pattern conversion, and DressCode's \cite{he2024dresscode} GPT-based text-conditioned pattern generation. These methods predict the panel information through its rotational matrix, translation matrix, vertex coordinates, edge connection relationships, edge types, and curvature, with stitches defined as a set of one-to-one connections between the edge of one panel and the edge of another panel. While effective for simple garments,  they struggle with complex designs due to the combinatorial explosion in pattern token sequences. Generated patterns often mismatch input, with issues such as self-intersections or incorrect stitching information, leading to simulation failures. Recent work mitigates these through large multimodal models \cite{bian2024chatgarment} and a novel tokenization scheme \cite{nakayama2024aipparel}. Our key insight diverges intrinsically: instead of directly predicting panel and stitch information, we focus on generating higher-level garment parameters, which are more intuitive elements and are easier to obtain from images.
% , and then use GarmentCode \cite{korosteleva2023garmentcode} to produce the corresponding sewing pattern from these parameters. 

\paragraph{Deformation-based 3D Garment Generation.} These methods typically take a pre-provided template garment as input and deform it to match the target garment using implicit fields and score distillation sampling (SDS). GarmentDreamer \cite{li2024garmentdreamer} and ClotheDreamer \cite{liu2024clothedreamer} employ text-guided deformation of 3D Gaussian Splatting (3DGS) representations. Garment3DGen \cite{sarafianos2024garment3dgen} introduces geometric supervision in 3D space by generating a coarse guidance mesh from image input and using it as a soft constraint during the deformation process. WordRobe \cite{srivastava2025wordrobe} designs a garment latent space and aligns it with the CLIP embedding space to enable text-guided garment generation and editing. DiffAvatar~\cite{li2023diffavatar} further optimize the garment to match the observation with proposed efficient differentiable simulation. While benefiting from recent advances in differentiable rendering and simulation, they still require lengthy optimization or deformation processes and exhibit strong dependency on the quality of the pre-provided template. Such limitations fundamentally restrict their adoption in real-time applications and large-scale production.

\paragraph{Autoregressive Models for 3D Generation.}
Auto-regressive transformers~\cite{vaswani2023attentionneed,MaskedAutoencoders2021} have radically transformed visual generation~\cite{esser2021taming,chang2022maskgit,muse23,var24,llamagen24,li2025autoregressive,chen2025mar,gu2025dart,ma2025tokenshuffle} through their sophisticated sequential approach of synthesizing images using discrete tokens derived from image tokenizers. This paradigm has achieved remarkable success by decomposing the complex task of image generation into a series of manageable token prediction steps, enabling more coherent and controllable outputs. 
% Building upon this foundation, recent work~\cite{meshanything,meshgpt,edgerunner} has introduced specialized mesh tokenizers that extend the auto-regressive framework to 3D mesh generation, providing high quality artist-created mesh. 
Building upon this foundation, recent works: SAR3D~\cite{sar3d} and MAR-3D~\cite{chen2025mar} effectively discretize 3D geometry into sequential tokens that can be predicted in an auto-regressive manner, similar to language modeling. More specifically, SAR3D developed a multi-scale quantization pipeline and uses VAR as the generation backbone. While MAR-3D combines the power of auto-regressive model and diffusion model to enable high-resolution mesh generation while maintaining computational efficiency. Our method builds upon these advances, first leveraging the strengths of auto-regressive models for garment parameters generation while using diffusion loss \cite{li2025autoregressive} that accounts for the continuous value space inherent in garment parameters, results in more realistic and physically plausible garment generations.

\section{Method}
\label{sec:method}

\begin{figure*}[t]
    \centering
    \includegraphics[width=\textwidth]{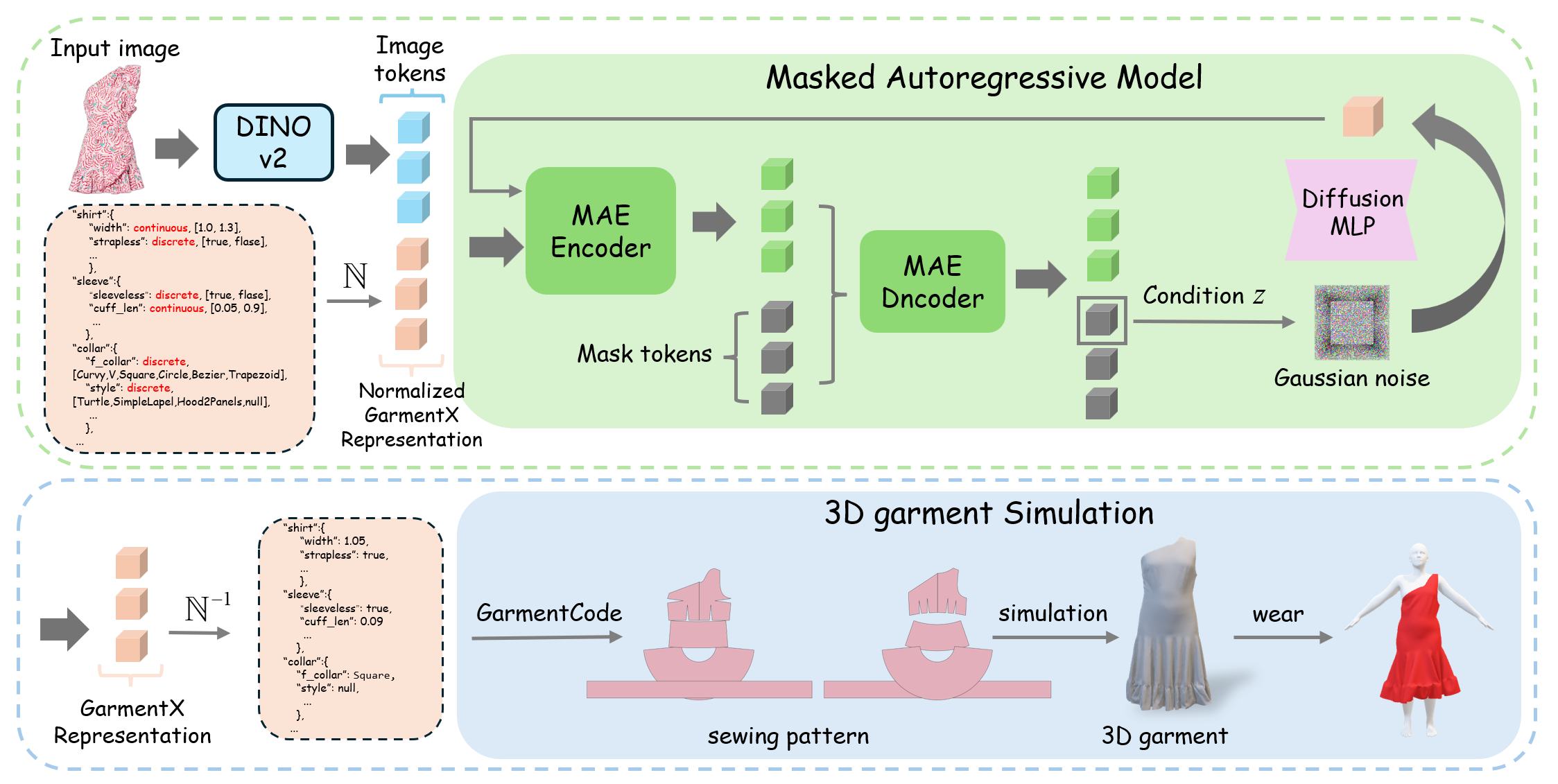}
	\caption{\textbf{Overview of \methodName{}.} Taking a single image as input, \methodName{} trains a masked autoregressive generation model directly upon our \methodName{} representation, extracts condition image tokens via DINOv2, processes them through MAE encoder-decoder architecture and diffusion MLP. The generated \methodName{} representation is projected back to original scale and then reconstruct sewing patterns through GarmentCode, and finally simulate 3D garments wearable on arbitrary human bodies.}
	\label{fig: garment_pipline}
\end{figure*}

We propose \methodName{}, a structured and parametric-driven framework for generating diverse, editable, and simulation-ready 3D garments from a single input image.
Central to our approach is a structured garment parameter representation that ensures physically valid and simulation-ready garment synthesis (Section~\ref{sec:rep}). To enable effective learning, we construct \methodName{} dataset, a large-scale dataset containing 378,682 garment parameter-image pairs, created through a novel automatic data construction pipeline based on the GarmentCodeData dataset (Section \ref{Data Pipline}). 
% To achieve this, we construct \methodName{} dataset, a large-scale dataset containing 378,682 garment parameter-image pairs, created through a novel automatic data construction pipeline by leveraging the publicly available GarmentCodeData dataset (Section \ref{Data Pipline}). 
% In addition, \methodName{} introduces a structured garment parameter representation that ensures simulation-ready garment generation (Section \ref{sec:rep}). 
% To achieve this, we designed a masked autoregressive (MAR) model to efficiently decode conditional image tokens into the editable garment parameters.
To generate structured garment parameters from an input image, we employ a masked autoregressive model to efficiently decode conditional image tokens into our representation.
Once generated, these garment parameters are transformed into 2D sewing patterns and simulated to 3D garments through GarmentCode (Section~\ref{Model Pipline}). We provide an overview of the proposed framework in Figure~\ref{fig: garment_pipline}.

% Furthermore, we meticulously designed a garment parameters normalization method that standardizes value ranges across different garments, combined with a MAR model to efficiently decode conditional image tokens into editable garment parameters. The generated garment parameters are finally simulated to 3D garments through GarmentCode (Sec. \ref{Model Pipline}). We provide an overview of the proposed framework in Figure \ref{fig: garment_pipline}.

% We propose \methodName{}, a framework to create diverse, open-structured, and simulation-ready 3D garments by generating garment parameters aligned with input images. To achieve this, we develop an automatic data construction pipeline to construct garment parameter-image pairs as training data based on the publicly available dataset GarmentCodeData (Sec. \ref{Data Pipline}). Furthermore, we meticulously designed a garment parameters normalization method that standardizes value ranges across different garments, combined with a MAR model to efficiently decode conditional image tokens into editable garment parameters. The generated garment parameters are finally simulated to 3D garments through GarmentCode (Sec. \ref{Model Pipline}). We provide an overview of the proposed framework in Figure \ref{fig: garment_pipline}.

\begin{figure}[t]
    \centering
    \includegraphics[width=\columnwidth]{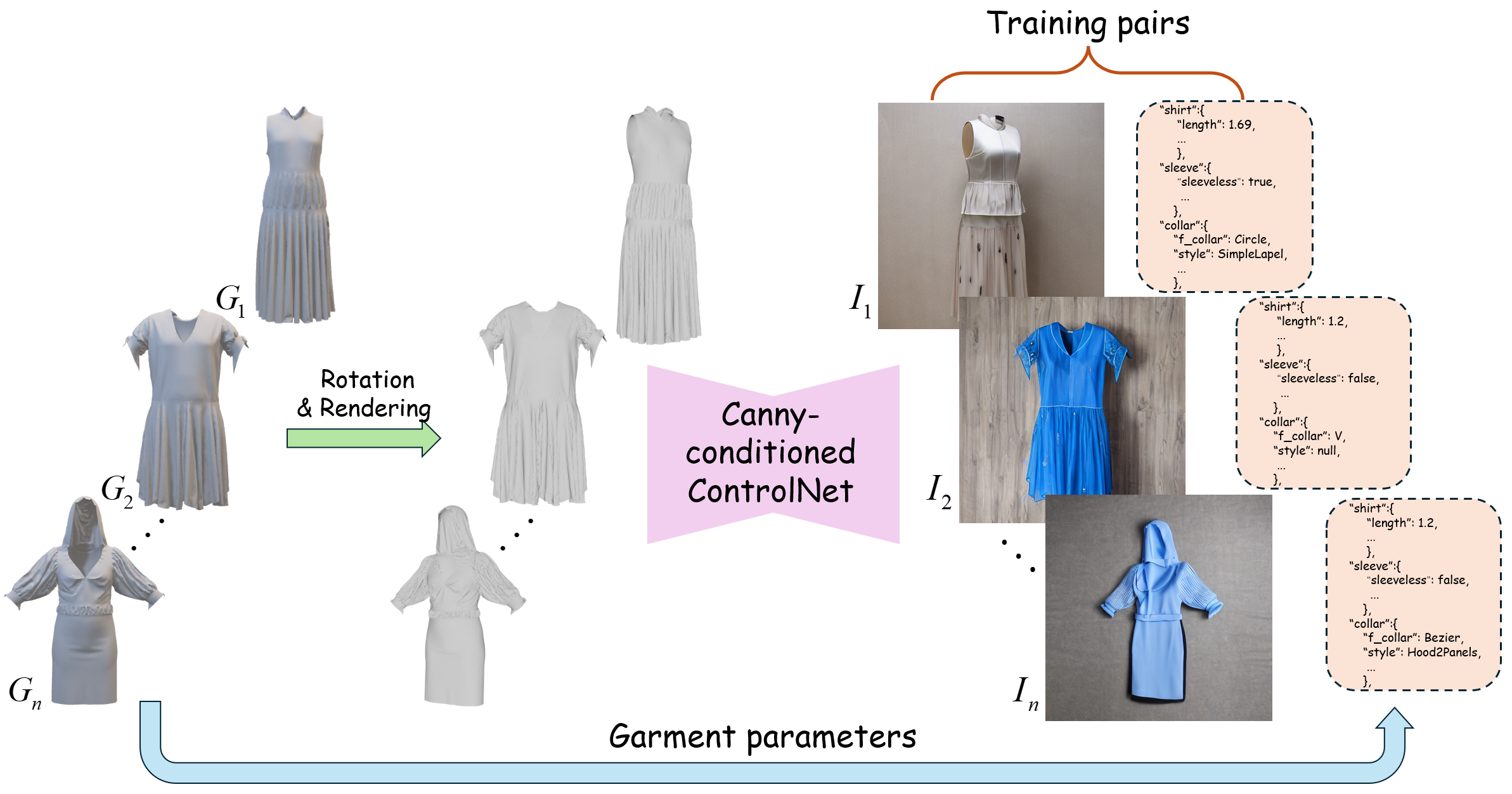}
	\caption{\textbf{Automatic Data Construction Pipeline.} We construct garment parameters-image pairs using ControlNet and Blender.}
	\label{fig: data_pipline}
\end{figure}

\subsection{GarmentX Parameter Representation}
\label{sec:rep}

A key innovation of \methodName{} is the introduction of a structured and editable parametric representation for garments. Instead of directly predicting 2D pattern edges and stitch relationships, which often results in geometrically invalid configurations, our approach represents garments using a structured parameter vector that can be decoded into a valid sewing pattern via GarmentCode~\cite{korosteleva2023garmentcode}.

Each garment is represented by a parameter vector $p = \left\{ {{p_0},{p_1}, \ldots ,{p_N}} \right\}$, where each ${p_i}$ corresponds to an editable garment attribute, such as neckline type, sleeve shape, pants length, and skirt flare.
This structured parameterization enables direct control over garment properties, facilitating both accurate synthesis and interactive editing.

To preserve relative positional relationships while ensuring numerical stability during diffusion training, the parameter values are normalized to a fixed range $\left[ { - 1,1} \right]$. For continuous parameters, normalization is performed using:

\begin{equation}
{x_i} = 2 \cdot \frac{{{p_i} - {p_{i\min }}}}{{{p_{i\max }} - {p_{i\min }}}} - 1,
\end{equation}

where $p_{i\min }$, $p_{i\max}$ and ${x_i}$ denote the minimum value, maximum value, and normalized value of the i-th parameter. 
For discrete attributes with $K$ ordinal choices, such as neckline type or sleeve style, we map each category to a continuous space uniformly partitioning $\left[ { - 1,1} \right]$ into $K$ intervals of width $\frac{2}{K}$.  
Each discrete option $k \in \left\{ {0,1, \ldots K} \right\}$ is encoded as the midpoint of its corresponding interval:
\begin{equation}
{x_i} = \frac{{2 \cdot \left( {k + 0.5} \right)}}{K} - 1
\end{equation}

For instance, when $K = 3$, the discrete choices $\left\{ {A,B,C} \right\}$ map to $\left\{ { - 0.666,0,0.666} \right\}$ respectively, creating a continuous space compatible with neural network operations.
The normalization operation can be expressed as $x = \mathbb{N}\left( p \right)$. During inference, the generated garment parameters are projected back to the original scale, represented as $p = {\mathbb{N}^{ - 1}}(x)$. The 3D garments are then generated using parameter $p$. 
This structured representation ensures consistent scaling across different garment attributes, improving training stability and generalization.

\subsection{\methodName{} Dataset Construction}
\label{Data Pipline}

A key requirement for training \methodName{} is a large-scale high-quality dataset containing paired garment images and our structured garment parameters.
However, existing datasets often lack fine-grained parametric annotations or are limited in garment diversity, restricting the generalization ability of learning-based methods.
To accommodate this need, we build \methodName{} dataset, a collection of 378,682 garment parameter-image pairs, using an automatic data construction pipeline that synthesizes realistc garment images aligned with \methodName{} parameter representation.

To ensure compatability with our framework, we transform the raw garment parameters from GarmentCodeData into our \methodName{} representation. 
GarmentCodeData provides low-level garment attributes, such as panel dimensions, edge connectivity, and stitch relationships, these parameters are often difficult to manipulate and prone to self-intersections when directly used for prediction.
By converting GarmentCodeData parameters into \methodName{} representations, we ensure that every sampled parameter vector is guaranteed to generate physically valid garments when decoded back into 2D sewing patterns and 3D simulations.

% A major challenge in training \methodName{} is the lack of large-scale high-quality paired data containing garment images and corresponding structured parameters. 
% Existing datasets often lack diversity in garment shapes and textures, limiting the generalization ability of learning-based methods. To address this, we introduce a novel automatic data construction pipeline that synthesizes high-quality garment images conditioned on parametric garment representation, forming the \methodName{} dataset.

% Training \methodName{} requires paired garment images and corresponding garment parameters. While GarmentCodeData provides 115,000 3D garments along with their garment parameters, but the images provided by GarmentCodeData have simple, single colors that exhibit significant domain gaps compared to real garment images. To bridge the data distribution discrepancy, we developed an automated data construction pipeline to generate photorealistic garment images while preserving parametric correspondence.

As shown in Figure \ref{fig: data_pipline}, given a set of 3D garments assets $G$, we render the multi-view images by randomly sampling camera poses from the combinations of three azimuths [0, 30, -30] and three elevations [0, 30, -30]. Once the multi-view images are generated, we enhance their realism using a canny-conditioned ControlNet \cite{zhang2023adding}, which transforms the rendered images into photorealistic counterparts. The base prompt ``A modern garment" ensures the generation of realistic, contemporary clothing styles, while supplementary descriptors such as ``best-quality," ``realistic," and ``photoreal" further enhance image fidelity.
Through this automated pipeline, we construct \methodName{} dataset, a large-scale collection containing 378,682 well-aligned garment parameter-image pairs.

% These rendered images then drive a canny-conditioned ControlNet \cite{zhang2023adding} to generate garment images $I$, guided by the fixed prompt: ``A modern garment". We add additional prompts (e.g. best-quality, realistic, photoreal,
% etc.) to ensure that ControlNet produces high-quality images. Finally, we obtain 378,682 aligned garment parameters-image pairs for training.

\subsection{3D Garment Generation}
\label{Model Pipline}

Our framework generates editable and physically valid 3D garments by predicting structured garment parameters through a masked autoregressive model. These parameters are then converted into sewing patterns and simulated into wearable 3D garments, ensuring realism, diversity, and controllability.

\noindent\textbf{Condition Scheme.}
We use DINOv2~\cite{oquab2023dinov2} to extract image features, which serve as initial tokens for the masked autoencoder (MAE \cite{he2022masked}) encoder. To enhance conditional generation quality, we apply classifier-free guidance (CFG), randomly nullifying conditional input features with a 0.1 probability during training~\cite{ho2022classifier}.

% For processing conditional images, we leverage DINOv2~\cite{oquab2023dinov2}. The images features serve as initial tokens for the MAR encoder, providing fine-grained features. During training, we implement classifier-free guidance by randomly nullifying conditional input features with 0.1 probability, enhancing conditional generation quality~\cite{ho2022classifier}.

\noindent\textbf{MAE Encoder and Decoder.}
The integration of image tokens and normalized GarmentX representations is achieved through concatenation, followed by random masking at a variable ratio of 0.7 to 1.0. In accordance with the MAE architectural principles, bidirectional attention mechanisms are implemented. The computational pipeline starts with the unmasked tokens through the MAE encoder, which comprises a series of self-attention layers. Subsequently, the encoded tokens are concatenated with the mask tokens with learnable positional embeddings.
% Prior to their introduction into the MAE decoder, both masked and unmasked tokens are augmented with learnable positional embeddings, thereby facilitating position-aware token prediction capabilities.

\noindent\textbf{Diffusion Process.}
For each token utilized in the diffusion process, the MAE decoder generates a corresponding condition vector $\mathbf{z} \in \mathbb{R}^D$. A MLP-based denoising network subsequently reconstructs ground truth tokens from Gaussian noise through the optimization of:

\begin{equation}
\mathcal{L}(z,x) = \mathbb{E}_{z,t}\left[|\epsilon - \epsilon_\theta(x^t|t,z)|^2\right]
\end{equation}
where $z$ represents the condition vector derived from the MAE decoder and $x_t$ denotes the ground truth parameters. During the inference phase, the reverse diffusion process is employed to predict each set of tokens in parallel.

\noindent\textbf{Inference Schedules.}
During inference, we first generate a random token generation order. Then we extract image tokens from input image, which are fed into the MAE encoder-decoder architecture. From the decoder outputs, we select condition vector $z$ according to the predetermined generation order. Multiple tokens undergo parallel DDIM~\cite{song2022ddim} denoising processes simultaneously. The number of tokens $N_s$ processed in each auto-regressive step follows a cosine schedule as in~\cite{chang2022maskgit}, progressively increasing over $S$ total steps:
\begin{equation}
\centering
\begin{aligned}
N_{s} = \left\lfloor N(cos(\frac{s}{S}) -cos(\frac{s-1}{S})) \right\rfloor
\end{aligned}
\label{eq:masking}
\end{equation}

This scheduling strategy is motivated by the observation that initial tokens are more challenging to predict, while later tokens become progressively easier to determine, similar to a completion task. Consequently, we generate fewer tokens in initial steps and gradually increase the number in later steps, rather than maintaining a constant generation rate across all steps.

\vspace{1mm}

\noindent\textbf{CFG Schedule.} We employ CFG in our diffusion model:
\begin{equation}
\centering
\begin{aligned}
\varepsilon = \varepsilon_{\theta}(x_t|t, z_u) + \omega_{s} \cdot (\varepsilon_{\theta}(x_t|t, z_c) - \varepsilon_{\theta}(x_t|t, z_u))
\end{aligned}
\label{eq:cfg}
\end{equation}
where $z_c$ and $z_u$ are conditional and unconditional output from the MAE decoder, which serve as the condition for the diffusion model. We employ a linear strategy~\cite{li2025autoregressive} for the CFG coefficient $\omega_{s}$, starting with lower values during the initial uncertain steps and progressively increasing it to $\lambda_{cfg}$. Specifically, in Eq.~\ref{eq:cfg}, we set $\omega_{s}=s\cdot\lambda_{cfg}/S$.

\noindent\textbf{3D Garment Simulation.}
Given the generated garment parameters, we leverage the established GarmentCode to first produce corresponding sewing patterns, followed by physically simulating 3D garments wearable on arbitrary human bodies. The strictly constrained parameter boundaries eliminate erroneous pattern generation observed in previous pattern-based approaches, thereby preventing simulation failures inherent to prior methods.

% \textbf{Model.} Given a condition image $I$, we use the pre-trained DINOv2 model to extract image features as tokens, denoted as $c = DINOv2\left( I \right)$. These tokens serve as conditional features and are fed into the DiT model $\epsilon$ to predict the noise for the noisy input ${{x_t}}$ at timestep $t$:
% \begin{equation}
% \epsilon\left( {{x_t},t,c} \right) = {\left\{ {CrossAttn\left( {SelfAttn\left( {{x_t},t} \right),c} \right)} \right\}^D}
% \end{equation} where $D$ denotes the number of DiT blocks in the model.

\section{Experiments}

\begin{table}[t]
    \centering
    \caption{\textbf{Quantitative comparisons on CLOTH3D dataset.} \methodName{} achieves the best performance.}
    \label{tab: Quantitative Comparisons}
    \renewcommand{\arraystretch}{1}%row space 
    \resizebox{\columnwidth}{!}{
        \begin{tabular}{cccccc}
		\hline
		\multicolumn{1}{c}{Method} & \multicolumn{1}{c}{CD $\downarrow$} & \multicolumn{1}{c}{P2S $\downarrow$} & \multicolumn{1}{c}{Failure Rate $\downarrow$} & \multicolumn{1}{c}{Runtime $\downarrow$} & \multicolumn{1}{c}{Wearable}\\
            \hline
		\multicolumn{1}{c}{SewFormer} & \multicolumn{1}{c}{$9.70$} & \multicolumn{1}{c}{$10.14$} & \multicolumn{1}{c}{76.3\%} & \multicolumn{1}{c}{$2$min} & 
        \multicolumn{1}{c}{\ding{51}}\\
		\multicolumn{1}{c}{LGM} & \multicolumn{1}{c}{$9.51$} & \multicolumn{1}{c}{$7.68$} & \multicolumn{1}{c}{-} & 
        \multicolumn{1}{c}{$82$s} & 
        \multicolumn{1}{c}{\ding{55}}\\
		\multicolumn{1}{c}{Garment3DGen} & \multicolumn{1}{c}{$6.34$} & \multicolumn{1}{c}{$5.91$} & \multicolumn{1}{c}{-} & \multicolumn{1}{c}{$15$min} & 
        \multicolumn{1}{c}{\ding{51}}\\
		\multicolumn{1}{c}{InstantMesh} & \multicolumn{1}{c}{$5.99$} & \multicolumn{1}{c}{$5.70$} & \multicolumn{1}{c}{-} & \multicolumn{1}{c}{$20$s} & 
        \multicolumn{1}{c}{\ding{55}}\\
		\multicolumn{1}{c}{Ours} & \multicolumn{1}{c}{\bm{$4.92$}} & \multicolumn{1}{c}{\bm{$4.96$}} & \multicolumn{1}{c}{\bm{$0$}} & \multicolumn{1}{c}{\bm{$15$}s} & 
        \multicolumn{1}{c}{\ding{51}}\\
		\hline
	\end{tabular}
    }
\end{table}

\subsection{Datasets, Metrics, and Implementation Details}

\textbf{Datasets.} We use \methodName{} dataset for training and the validation set of CLOTH3D \cite{bertiche2020cloth3d} for qualitative and quantitative comparison with baselines.
% We utilize GarmentCodeData \cite{korosteleva2024garmentcodedata} combined with our automatic data construction pipeline to obtain the training data pairs. For validation, we chose the validation set of CLOTH3D \cite{bertiche2020cloth3d} for qualitative and quantitative comparison with baselines, as it provides detailed 3D garment meshes, materials, and textures. 
Due to the limited garment styles in CLOTH3D, especially for fashion dresses, we supplemented our qualitative evaluation with the recently proposed Complex Virtual Dressing Dataset (CVDD) \cite{jiang2024fitdit}, which contains diverse and intricate fashion garments.

\noindent\textbf{Metrics.} We employ Chamfer Distance (CD), Point-to-Surface distance (P2S), simulation failure rate, and runtime as quantitative metrics. We also record whether each method produces wearable results to evaluate its suitability for downstream tasks.

% Some pattern-based methods may produce predicted patterns with self-intersections, incorrect stitch information, or undetectable panels, leading to simulation failure. When calculating CD and P2S, we filter out these failure cases and only consider successful simulations.

\begin{figure*}[t]
    \centering
    \includegraphics[width=\textwidth]{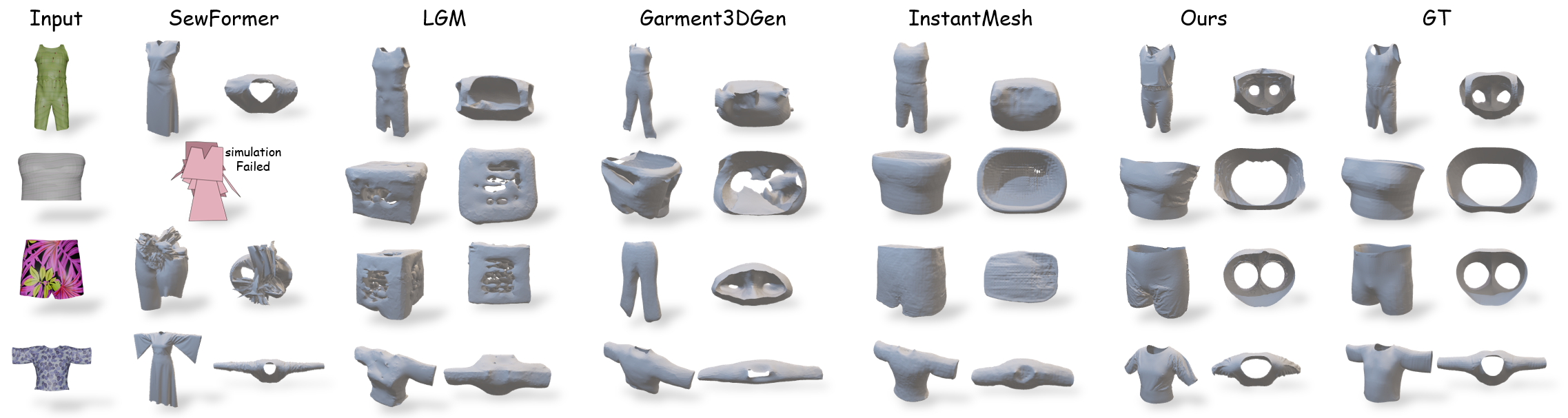}
	\caption{\textbf{Qualitative comparisons on CLOTH3D dataset.} \methodName{} produce wearable, open-structure, and complete 3D garments.}
	\label{fig: comp_CLOTH3D}
\end{figure*}

% \subsection{Implementation Details}
\noindent\textbf{Implementation Details.} Input images are resized into $224 \times 224$ and processed by the DINOv2 ViT-B/14 model to extract condition tokens. Training completes in 50 hours on two GPUs with a batch size of 48. More details are provided in supplementary materials.

% We train the model using the AdamW optimizer (${\beta _1}$=0.9 and ${\beta _2}$=0.999) with an initial learning rate of $1e - 4$ and a weight decay of $0.01$. The learning rate follows a cosine annealing schedule with warm restarts, starting with an initial cycle length of 10 epochs and decaying to 1\% of the initial rate. 

\subsection{Comparisons}
We evaluate our approach against three types of methods: 1) pattern-based methods, including SewFormer \cite{liu2023towards} and DressCode \cite{he2024dresscode}, which directly predict the panel and stitch information of sewing pattern; 2) deformation-based methods, such as Garment3DGen \cite{sarafianos2024garment3dgen}; and 3) general 3D generation methods, including InstantMesh \cite{xu2024instantmesh}, LGM \cite{tang2025lgm} and Trellis \cite{xiang2024structured}. Since DressCode is a text-guided method, to ensure fairness, we compare it only from the perspective of generalization performance.

% Unfortunately, since GarmentDreamer \cite{li2024garmentdreamer}, ClotheDreamer \cite{liu2024clothedreamer}, and WordRobe \cite{srivastava2025wordrobe} did not provide their code, we cannot compare our method with them. 

\noindent\textbf{Quantitative Comparisons.} Table \ref{tab: Quantitative Comparisons} presents quantitative comparisons between our method and baseline approaches. Our method performs best on both the CD and P2S metrics, surpassing the strongest competitor InstantMesh by margins of 1.07 in CD and 0.74 in P2S metrics. This demonstrates its superior capability in accurately generating 3D garments aligned with input images. Previous direct sewing pattern prediction approaches like SewFormer exhibit fundamental limitations—their 76.3\% simulation failure rate stems from frequent self-intersection panels and erroneous stitch predictions. Our method fundamentally resolves these issues, achieving zero failure cases. Additionally, our method maintains efficient inference time of 15 seconds per generation, a 60$\times$ speed improvement over deformation-based methods such as Garment3DGen, which takes 15 minutes per sample.

\begin{figure}[t]
    \centering
    \includegraphics[width=\columnwidth]{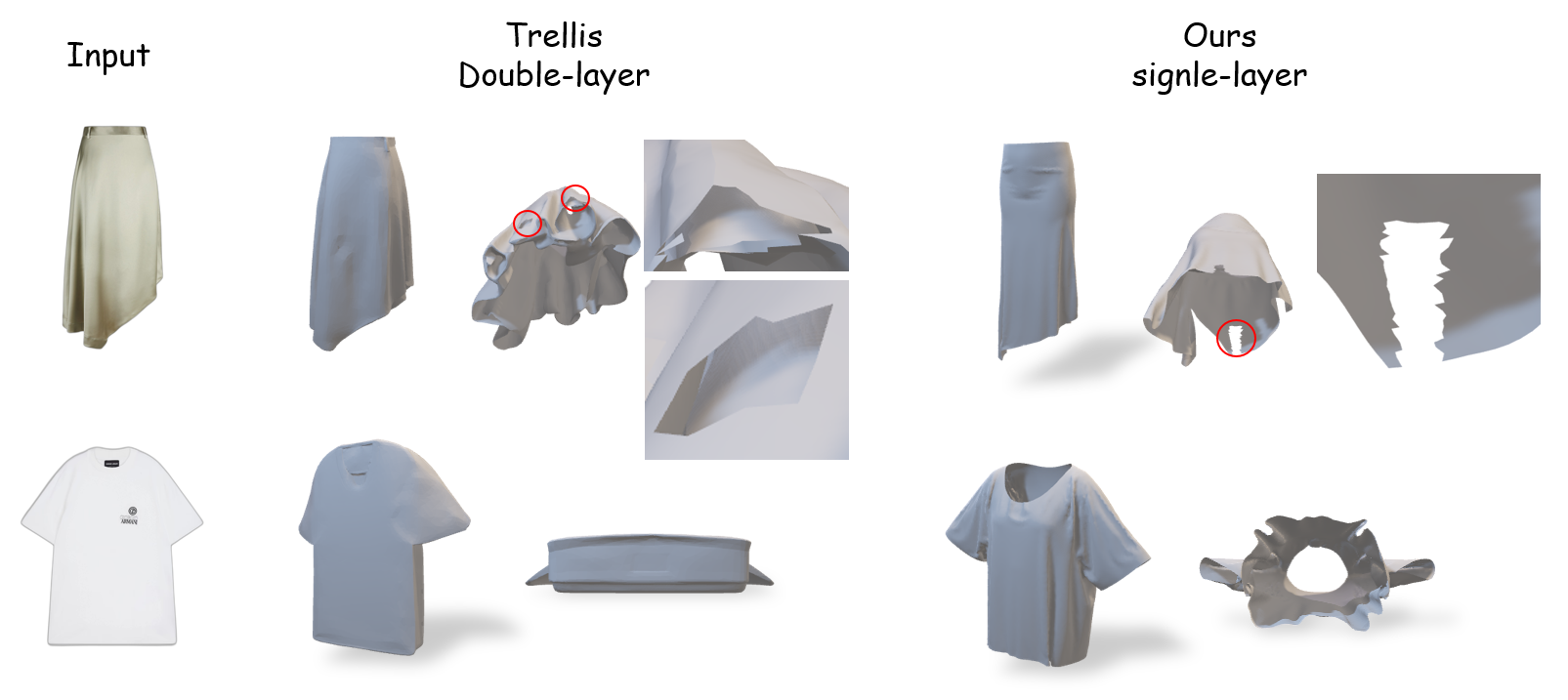}
	\caption{\textbf{Qualitative comparisons with Trellis.} \methodName{} generates single-layer, open-structure 3D garments.}
	\label{fig: comp_Trellis}
\end{figure}

\noindent\textbf{Wearable.} While InstantMesh and LGM are competitive in quantitative metrics, they tend to produce closed garments that cannot be worn directly, as evidenced by the side and top views in Figure \ref{fig: comp_CLOTH3D}. Furthermore, as evidenced by cross-sectional analysis through mesh cutting operations in MeshLab, shown in Figure \ref{fig: comp_Trellis}, Trellis generates double-layer meshes with some sticky areas at the ends of skirts, and it completely closes when dealing with the T-shirt images. In contrast, our method produces single-layer, open-structure garments, which can be directly draped onto various human bodies and seamlessly animated through simulators.

\begin{figure*}[t]
    \centering
    \includegraphics[width=\textwidth]{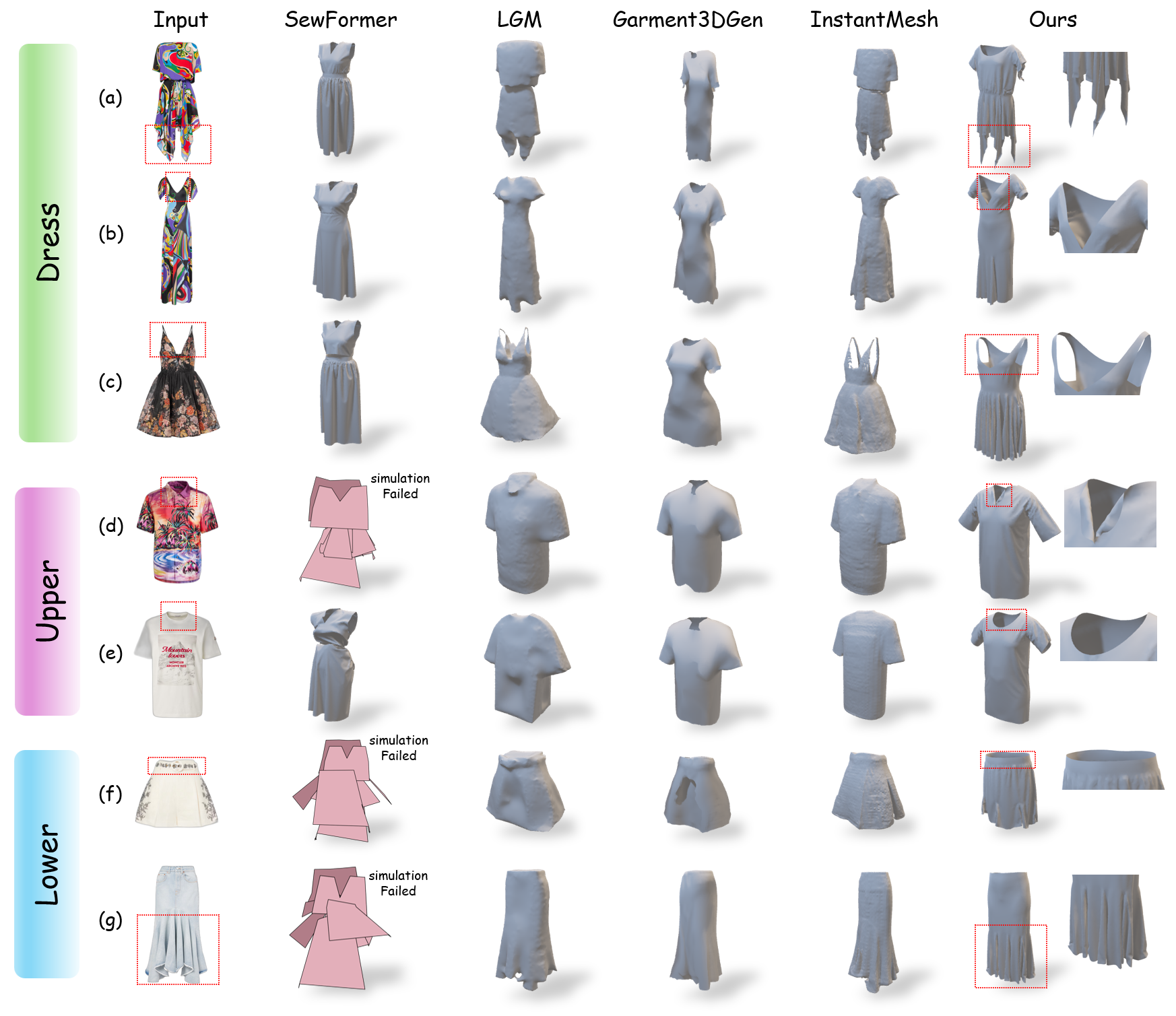}
	\caption{\textbf{Qualitative comparisons on CVDD dataset.} \methodName{} produce diverse, complex, and detailed 3D garments that match the input images well.}
	\label{fig: comp_CVDD}
\end{figure*}

\noindent\textbf{Qualitative Comparisons.} Figure \ref{fig: comp_CLOTH3D}, \ref{fig: comp_Trellis}, and \ref{fig: comp_CVDD} provide qualitative comparisons between our method and baseline approaches. SewFormer exhibits limited generalization capability beyond the training data distribution, thereby constraining reconstructable garment diversity, and frequently resulting in geometrically distorted garments due to flawed stitches. Garment3DGen's performance is heavily dependent on pre-provided 3D garment templates, leading to highly mismatched or even incomplete results when the template and input image differ substantially. In contrast, our method demonstrates precision in reconstructing nuanced features, as shown in Figure \ref{fig: comp_CVDD}, such as irregular ribbons (a, g), collar (b, d, e), shoulder belt (c), and waistline (f), while the baseline methods fail to reproduce these details. More results are included in the supplementary materials.

\begin{figure}[t]
    \centering
    \includegraphics[width=0.9\columnwidth]{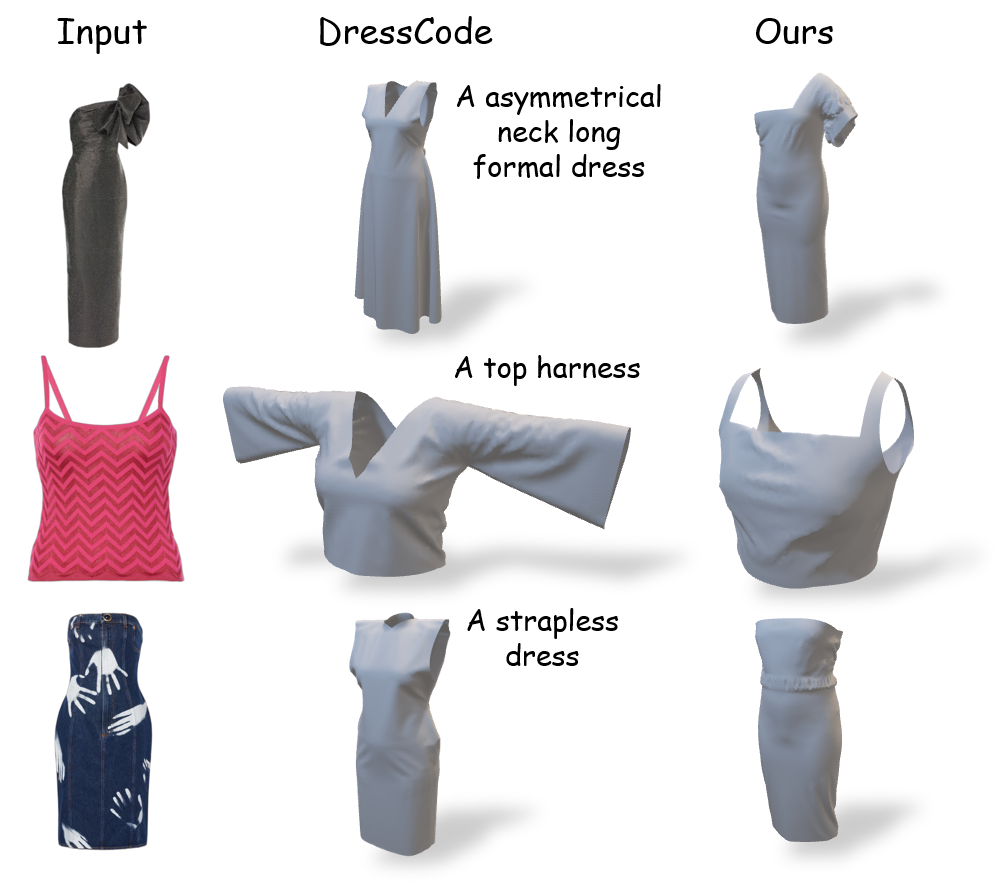}
	\caption{\textbf{Qualitative comparisons with DressCode.} \methodName{} exhibits better generalization, enabling the creation of a variety of uncommon garments.}
	\label{fig: comp_DressCode}
\end{figure}

\begin{figure}[t]
    \centering
    \includegraphics[width=0.9\columnwidth]{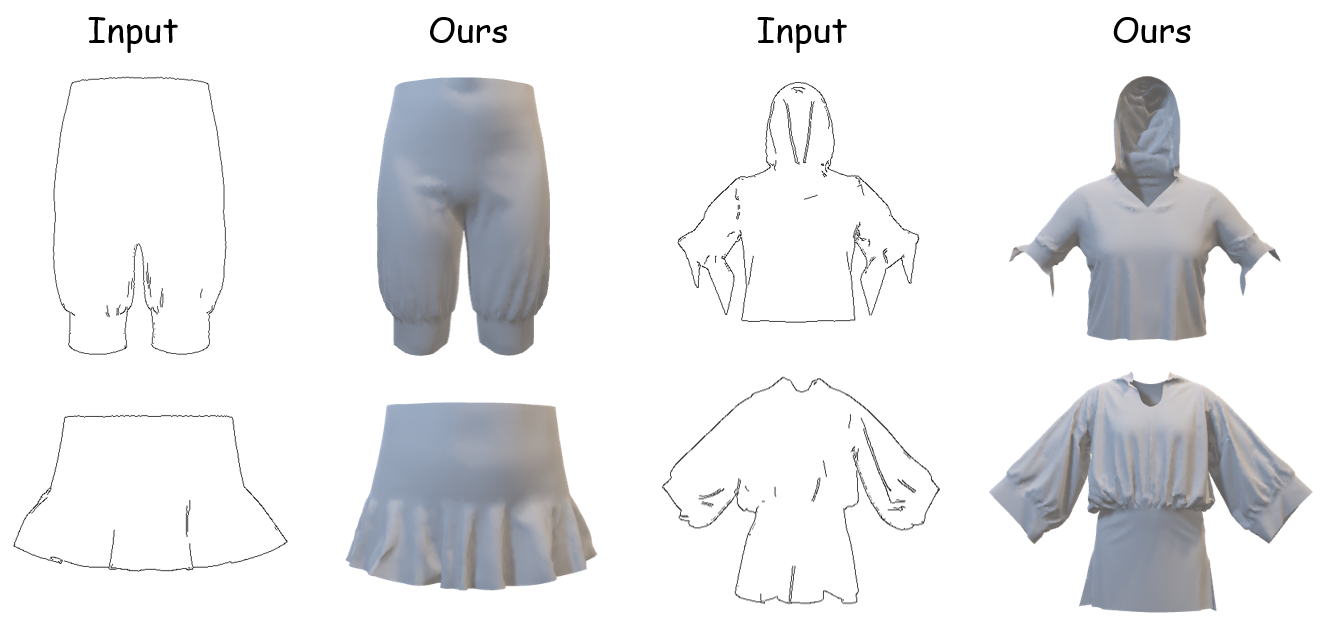}
	\caption{\textbf{Sketch-conditional inputs.} Note that \methodName{} has not been trained on sketch images.}
	\label{fig: comp_Sketch}
\end{figure}

\noindent\textbf{Generalization.} We also evaluate the generalization capabilities of our method in comparison to DressCode, with comparative results visualized in Figure \ref{fig: comp_DressCode}. Our approach demonstrates superior performance in generating diverse fashion garments, such as asymmetric one-shoulder dresses, top harnesses, and strapless dresses, which DressCode fails to produce. Additionally, as evidenced in Figure \ref{fig: comp_Sketch}, despite never being trained on sketch images, our model consistently generates 3D garments that maintain precise geometric alignment with the input sketch contours. 

% These results collectively highlight the exceptional generalization performance of our method across different input modalities and garment types, even when operating outside the distribution of the training set.

\subsection{Ablation Study}
\begin{figure}[t]
    \centering
    \includegraphics[width=0.9\columnwidth]{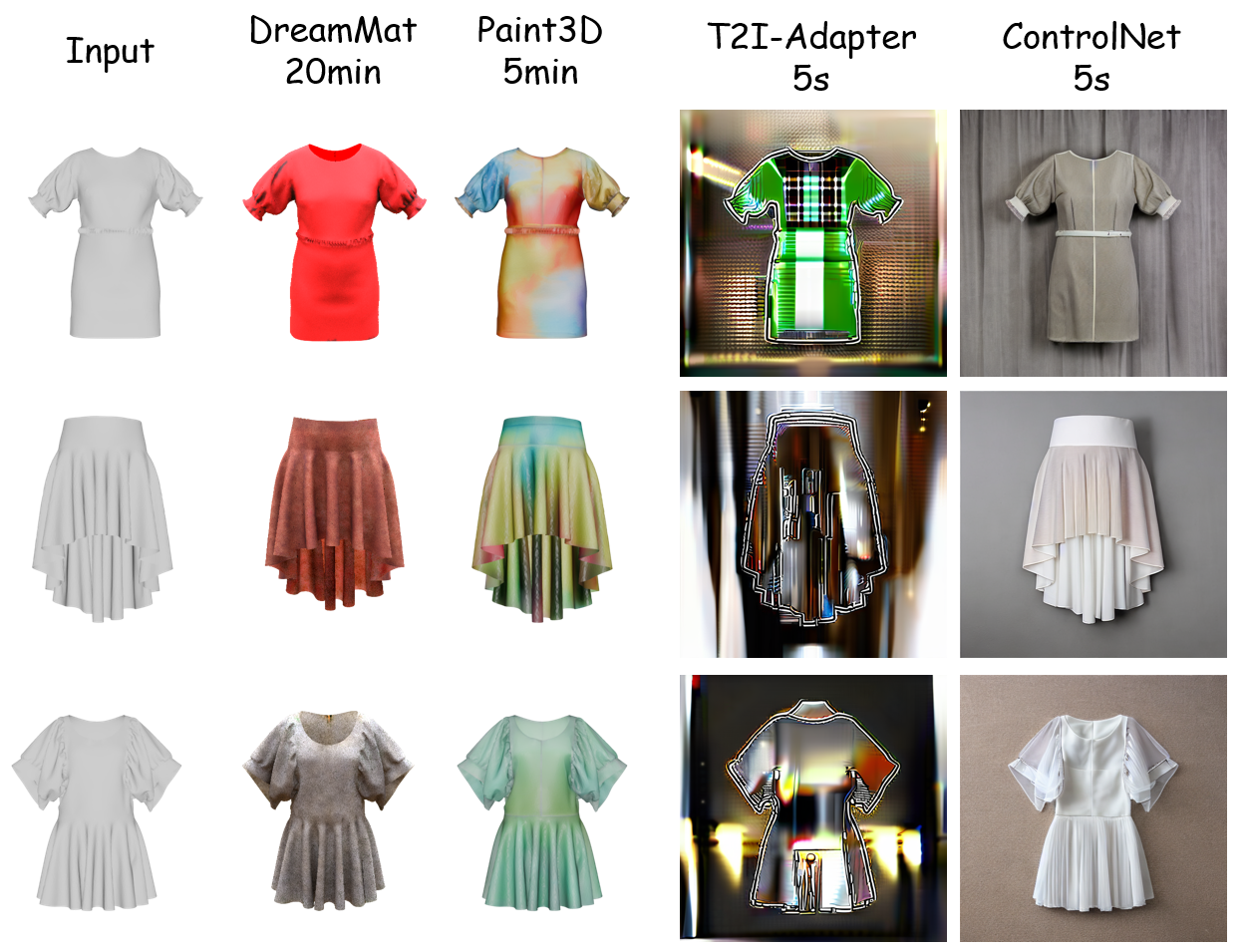}
	\caption{\textbf{Qualitative ablation of different data construction methods.} ControlNet was selected for its optimal balance between speed and quality.}
	\label{fig: comp_data}
\end{figure}

\textbf{Choice of Data Construction Methods.} There are two possible approaches for obtaining a garment image from a 3D garment mesh: render first and then generate the image, or generate textures first and then render it. For the first approach, we compared two canny-conditional Stable Diffusion models: ControlNet and T2I-Adapter \cite{mou2024t2i}. For the second approach, we generated textures with Paint3D \cite{zeng2024paint3d} or DreamMat \cite{zhang2024dreammat} and then rendered them. We chose the first method because texture generation takes 5 minutes with Paint3D and 20 minutes with DreamMat, making it difficult to conduct large-scale data construction, whereas ControlNet or T2I-Adapter can generate images in 5 seconds. We ultimately select ControlNet due to T2I-Adapter producing lower quality images that often failed to resemble garments, as shown in Figure \ref{fig: comp_data}.

\noindent\textbf{Autoregressive model vs DiT \cite{peebles2023scalable}.} To verify the accuracy of our autoregressor in generating garment parameters, we conducted experiments by replacing our model with DiT. As quantitatively demonstrated in Table \ref{tab: model ablation}, our method achieves a 1.25 lower CD metric and a 1.24 lower P2S metric compared to DiT-XL, while requiring 43\% fewer parameters. This advantage stems from our autoregressive formulation that decomposes complex joint distribution modeling into a product of simpler marginal distributions, enabling faster parallel generation and higher quality outputs.

\noindent\textbf{Ablation on CFG Scale.} To balance the generation quality and condition strength, we systematically evaluate CFG scale ${\lambda _{cfg}} \in \left[ {2.0,5.0} \right]$. As Table~\ref{tab: CFG scaling factors ablation} demonstrates, $\lambda_{cfg} = 3.0$ achieves the optimal trade-off, which is consequently adopted in our model.

\begin{table}[t]
    \centering
    \caption{\textbf{Quantitative ablation of model}. Our autoregressor outperforms DiT on CLOTH3D dataset and also uses a more optimal number of parameters.}
    \label{tab: model ablation}
    \renewcommand{\arraystretch}{1}%row space 
    \resizebox{0.8\columnwidth}{!}{
        \begin{tabular}{cccc}
		\hline
		\multicolumn{1}{c}{Method} & \multicolumn{1}{c}{CD $\downarrow$} & \multicolumn{1}{c}{P2S $\downarrow$} & \multicolumn{1}{c}{num parameters} \\
            \hline
		\multicolumn{1}{c}{DiT-S} & \multicolumn{1}{c}{$6.47$} & \multicolumn{1}{c}{$6.40$} & \multicolumn{1}{c}{$30$M}\\
		\multicolumn{1}{c}{DiT-B} & \multicolumn{1}{c}{$6.47$} & \multicolumn{1}{c}{$6.35$} & \multicolumn{1}{c}{$119$M}\\
		\multicolumn{1}{c}{DiT-XL} & \multicolumn{1}{c}{$6.17$} & \multicolumn{1}{c}{$6.20$} & \multicolumn{1}{c}{$607$M}\\
		\multicolumn{1}{c}{Ours} & \multicolumn{1}{c}{\bm{$4.92$}} & \multicolumn{1}{c}{\bm{$4.96$}} & \multicolumn{1}{c}{$352$M}\\
		\hline
	\end{tabular}
    }
\end{table}

\begin{table}[t]
    \centering
    \caption{\textbf{Quantitative compartion of CFG scaling factor on CLOTH3D dataset.} We finally adopt 3.0 in our model.}
    \label{tab: CFG scaling factors ablation}
    \renewcommand{\arraystretch}{1}%row space 
    \resizebox{0.5\columnwidth}{!}{
        \begin{tabular}{cccc}
		\hline
		\multicolumn{1}{c}{CFG scale} & \multicolumn{1}{c}{CD $\downarrow$} & \multicolumn{1}{c}{P2S $\downarrow$}\\
            \hline
		\multicolumn{1}{c}{2.0} & \multicolumn{1}{c}{$5.14$} & \multicolumn{1}{c}{$5.21$} \\
		\multicolumn{1}{c}{3.0} & \multicolumn{1}{c}{\bm{$4.92$}} & \multicolumn{1}{c}{\bm{$4.96$}} \\
		\multicolumn{1}{c}{4.0} & \multicolumn{1}{c}{$5.37$} & \multicolumn{1}{c}{$5.38$} \\
		\multicolumn{1}{c}{5.0} & \multicolumn{1}{c}{$5.03$} & \multicolumn{1}{c}{$5.07$} \\
		\hline
	\end{tabular}
    }
\end{table}

\subsection{Garment Editing}
The parametric representation framework inherently facilitates geometry-aware garment customization. As shown in the lower panel of Figure \ref{fig: overall}, our generated garment parameters enable intuitive modification of style features (e.g., removing sleeves), components design (e.g., adding a hood), and even garment category (e.g., transforming dress into a jumpsuit) through minimal parameter adjustments. This parametric representation framework significantly enhances interoperability with downstream applications such as fashion design platforms.

\section{Conclusion and Future work}
We introduce \methodName{}, a structured and parametric-driven framework for editable and physically valid 3D garment generation. By leveraging a masked autoregressive model, \methodName{} efficiently predicts structured garment parameters, ensuring high-fidelity garment synthesis. Additionally, we construct \methodName{} dataset, a large-scale data collection of garment-parameter-image pairs, enabling robust training and improved generalization.

% This paper presents a pattern-based 3D garment generation framework that creates 3D garments through image-guided garment parametric modeling. Unlike direct pattern prediction methods, our system first generates adjustable garment parameters, then decodes them into reasonable sewing patterns via GarmentCode, followed by physics-based simulation. This approach fundamentally eliminates simulation failures caused by flawed sewing pattern predicted by prior works. The proposed MAR model achieves rapid parameter generation versus deformation-based methods requiring 15min optimization. Extensive experiments demonstrate our method's capability in producing diverse, open-structure garments ready for downstream tasks.

\methodName{} shows significant advances but has limitations. It is currently trained on human-free garment images and requires such images as input. For photos with wearers, we adopt SAM2 \cite{ravi2024sam} to segment clothing before generation (results are included in supplementary materials). Future work will expand the proposed automatic data construction pipeline to clothed humans, enabling direct processing without segmentation.

{
    \small
    \bibliographystyle{ieeenat_fullname}
    \bibliography{main}
}

\maketitlesupplementary

\noindent In the supplementary material, we include additional details
on training and evaluation, as well as more qualitative visualizations, future works, and failure cases.

\section{More Implementation Details}
The MAR component adopts a Transformer-based encoder-decoder architecture, featuring 16 layers in both the encoder and decoder paths, 16 attention heads per layer, and an embedding dimension of 768. This design choice facilitates effective modeling of long-range dependencies in 3D geometry. The MLP-based diffusion module consists of 6 layers with a width of 1024, which provides sufficient capacity for the denoising process while maintaining computational efficiency. We train our model using the AdamW optimizer with an initial learning rate of $1e - 4$, ${\beta _1}$=0.9, ${\beta _2}$=0.999, $\varepsilon  = 1 \times {10^{ - 8}}$, and a weight decay of $0.01$. The learning rate follows a cosine annealing schedule with warm restarts, starting with an initial cycle length of 10 epochs and decaying to 1\% of the initial rate. This scheduling strategy helps stabilize early training while ensuring convergence in later stages. During inference, the reverse denoising steps are set to 100, and the auto-regressive steps are set to 32 for MAR.

\section{More Qualitative Results}
\begin{figure*}[h]
    \centering
    \includegraphics[width=0.88\textwidth]{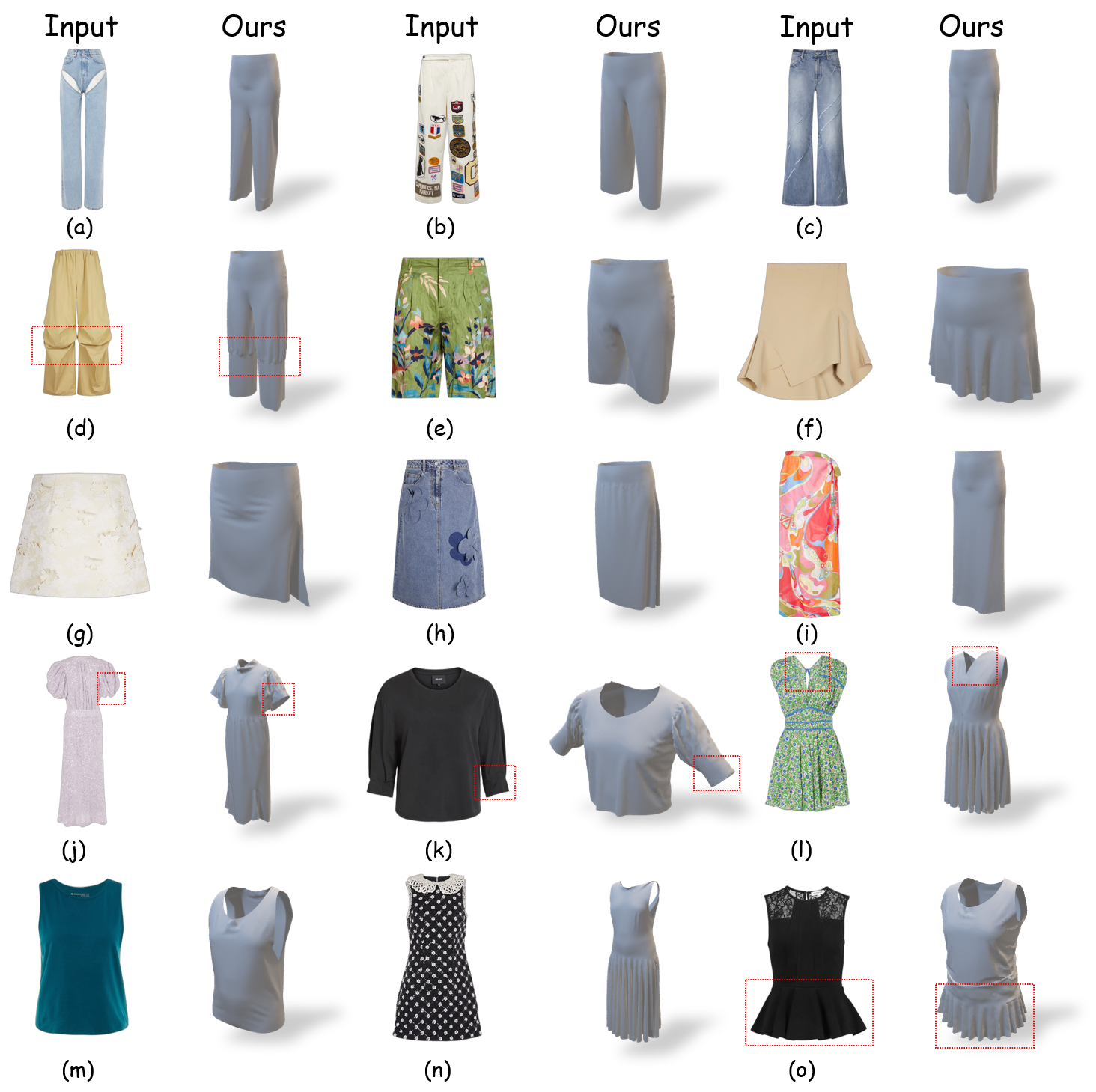}
	\caption{\textbf{More qualitative results.} \methodName{} can handle diverse input images with various styles and textures. It accurately captures geometric details in the input images and generates high fidelity, open-structure and simulation-ready 3D garments, facilitating downstream applications.}
	\label{fig: more result}
\end{figure*}

Here we show more qualitative results of \methodName{}. As shown in the first three rows of Figure~\ref{fig: more result}, our framework successfully generates various types of lower garments, accurately distinguishing between full-length pants (a, b, c, d), shorts (e), miniskirts (f, g), and bodycon skirts (h, i). The subsequent rows showcase generated dresses and upper garments, our method captures intricate geometric details including fitted cuffs (j, k), collars (l), sleeveless design (l, m, n, o), and pleated skirt design (o).

\section{Future Works}
\begin{figure}[h]
    \centering
    \includegraphics[width=0.9\columnwidth]{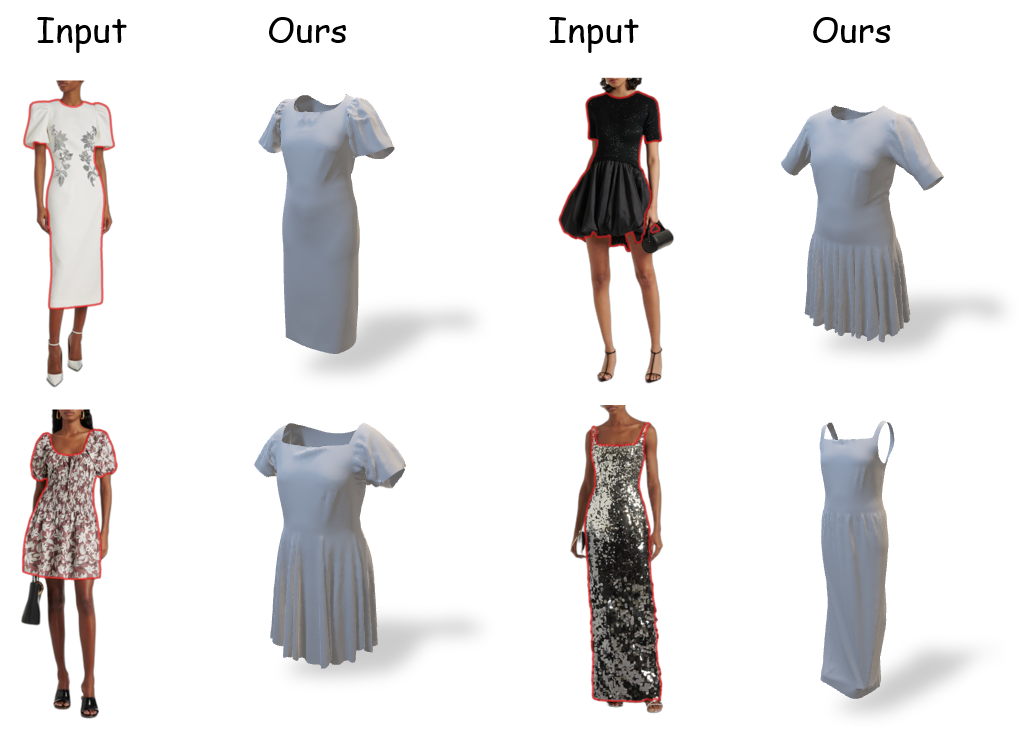}
	\caption{\textbf{Segmentation result as Input.} We use SAM2 to get the segmentation result and use it as the input to our model.}
	\label{fig: sam result}
\end{figure}

The current \methodName{} implementation processes only human-free garment images due to our automatic data generation pipeline's human-free design. For clothed human inputs as shown in Figure~\ref{fig: sam result}, we employ SAM2 to segment for preprocessing. Planned improvements include expanding the automatic data construction pipeline to encompass clothed-human images, thereby enabling end-to-end processing of clothed-human images.

\section{Failure Cases}
\begin{figure*}[h]
    \centering
    \includegraphics[width=0.85\textwidth]{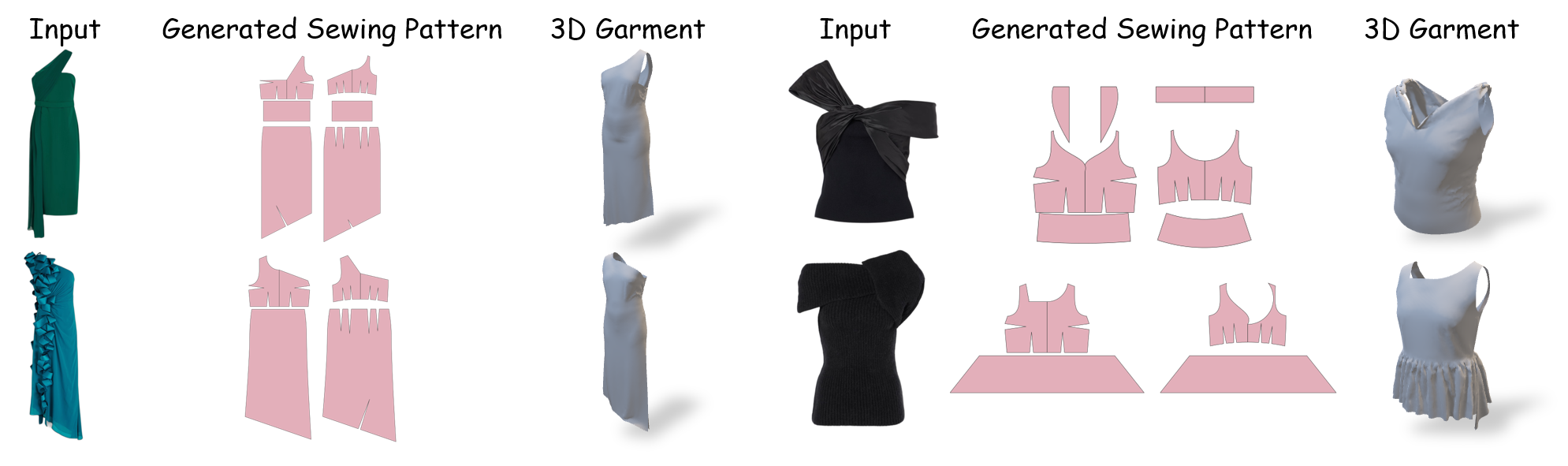}
	\caption{\textbf{Some failure cases.} \methodName{} fails when handling highly complex cases.}
	\label{fig: failed case}
\end{figure*}

While \methodName{} supports a wide range of modern garment categories, Figure~\ref{fig: failed case} reveals failure cases in processing highly complex designs—particularly those containing complex decorations, like oversized chest-attached bow knots and petal-like designs. For these challenging cases where even professional artists would struggle to conceptualize corresponding sewing patterns, our system nevertheless generates the closest valid parameter set and sewing patterns to approximate the images. This contrasts with direct sewing pattern prediction methods that often yield simulating failed patterns under such complexity.

\end{document}